\documentclass[11pt]{article}

\usepackage{acl}

\usepackage{times}
\usepackage{latexsym}

\usepackage[T1]{fontenc}

\usepackage[utf8]{inputenc}

\usepackage{microtype}

\usepackage{inconsolata}

\usepackage{graphicx}

\usepackage{amsmath}
\usepackage{amsfonts}
\usepackage{booktabs}
\usepackage{multirow}
\usepackage{dsfont}
\usepackage{multirow}
\usepackage{hhline}
\usepackage{tabularray}
\usepackage{array}
\usepackage{graphicx}
\usepackage{makecell}
\usepackage{comment}
\usepackage{xcolor}
\usepackage[export]{adjustbox}
\usepackage{subcaption}
\usepackage{tabularx,listings}
\usepackage{longtable}

\title{Region-R1: Reinforcing Query-Side Region Cropping for Multi-Modal Re-Ranking}

\author{Chan-Wei Hu, Zhengzhong Tu \\
  Texas A\&M University \\
  \texttt{\{huchanwei123,tzz\}@tamu.edu} \\
  }

\begin{document}
\maketitle

\begin{abstract}
Multi-modal retrieval-augmented generation (MM-RAG) relies heavily on re-rankers to surface the most relevant evidence for image-question queries. 
However, standard re-rankers typically process the full query image as a global embedding, making them susceptible to visual distractors (e.g., background clutter) that skew similarity scores.
We propose \textbf{Region-R1}, a query-side region cropping framework that formulates region selection as a decision-making problem during re-ranking, allowing the system to learn to retain the full image or focus only on a question-relevant region before scoring the retrieved candidates. Region-R1 learns a policy with a novel region-aware group relative policy optimization (r-GRPO) to dynamically crop a discriminative region. Across two challenging benchmarks, E-VQA and InfoSeek, Region-R1 delivers consistent gains, achieving state-of-the-art performances by increasing conditional Recall@1 by up to 20\%. These results show the great promise of query-side adaptation as a simple but effective way to strengthen MM-RAG re-ranking.
\end{abstract}

\begin{figure*}[ht]
  \centering
  \includegraphics[width=6.3in]{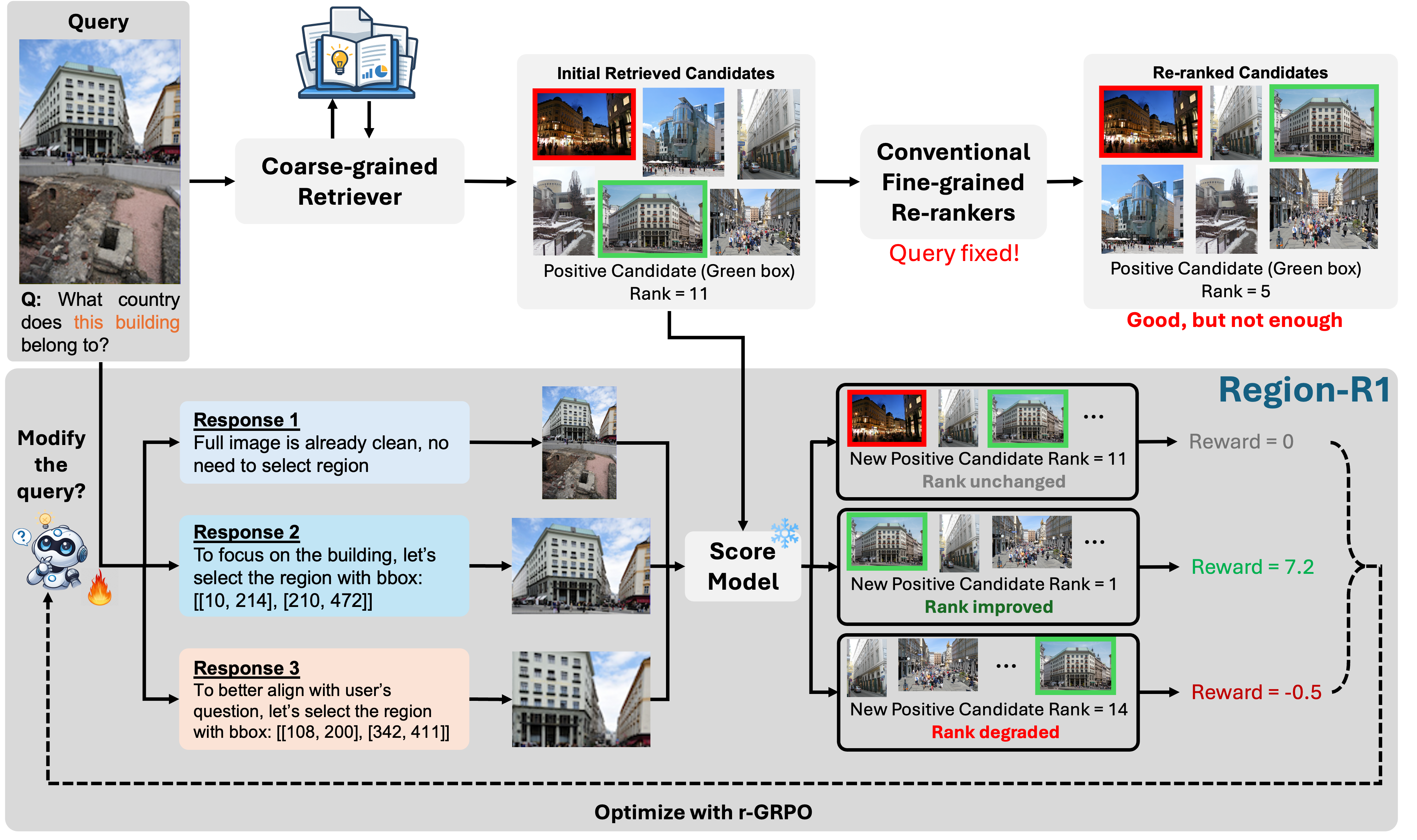}
  \caption{\textbf{Overview of query-side region cropping approach.} Conventional re-rankers treat the query as fixed and only re-order candidates. Our method instead automatically adapt the query by selecting an informative region or keeping the original query before scoring candidates.}
  \label{fig:overview}
\end{figure*}

\section{Introduction}
\label{sec:intro}

Multi-modal retrieval-augmented generation (MM-RAG) has become an increasingly important paradigm for grounding vision language models (VLMs) with both textual and visual information~\cite{reveal,lin2022retrieval,10.5555/3618408.3620067, murag}.
In MM-RAG systems, queries may include images in addition to natural language, and retrieved evidence may consist of heterogeneous modalities such as images, captions, diagrams, or structured visual content.
As a result, the effectiveness of MM-RAG depends not only on language understanding, but also on how visual information is represented.

In practical MM-RAG pipelines, an upstream retriever first produces a candidate set, after which a re-ranking model refines their ordering with respect to a given query.
Recent work has primarily focused on improving retrievers~\cite{flmr, lin2022retrieval,marvel} or designing more expressive multi-modal re-rankers~\cite{yu2024rankrag,echosight,omgm}.
By contrast, comparatively little attention has been paid to how the \emph{multi-modal query representation itself} is constructed and controlled during re-ranking.
In particular, re-rankers typically rely on a single global embedding of the query image, implicitly assuming that all regions of the image are relevant to the user’s question.

This assumption is often violated in realistic MM-RAG scenarios.
Query images frequently contain distractive objects, background regions, or contextual elements that are irrelevant to the question being asked.
When such irrelevant regions dominate the global visual representation, they can distort visual similarity estimates and degrade re-ranking performance, even when the candidate set is fixed.
This issue is specific to multi-modal settings and does not arise in purely textual RAG.

A natural response is to perform region cropping on the query image in a question-conditioned manner.
Modern vision-language models exhibit strong localization capabilities, suggesting that they can identify regions relevant to a given question.
Indeed, our preliminary analysis shows that, under a fixed candidate pool and a scoring model, replacing the full query image with an appropriately selected region can substantially improve multi-modal re-ranking performance.
However, naive or heuristic cropping strategies can also remove useful visual context and lead to worse rankings.
This raises a key challenge for MM-RAG systems: \emph{how can query-side visual information be selectively modified in a way that reliably improves multi-modal re-ranking performance?}

In this work, we formulate query-side region cropping as a decision problem optimized directly for multi-modal re-ranking objectives, named \textbf{Region-R1}.
We focus on the re-ranking stage because retrieval is a coarse-grained, high-recall step that must run efficiently at corpus scale, whereas re-ranking performs fine-grained discrimination over a small top-$K$ set.
Applying region cropping during retrieval would require policy execution during large-scale negative mining and may even necessitate re-indexing, substantially increasing training and system cost. In addition, it also risks discarding information too early and hurting recall.
Accordingly, we apply region cropping only at re-ranking, where computation is bounded and the selected region helps disambiguate among already plausible candidates.

Given a query image and user question, a policy determines whether to retain the full image or to select a region, with the explicit goal of improving the ranking of a fixed multi-modal candidate set.
Our model learns the region cropping policy using reinforcement learning, with rewards defined as improvements over baseline measured by standard information retrieval metrics such as Mean Reciprocal Rank~\cite{mrr} and Normalized Discounted Cumulative Gain~\cite{ndcg}.


We evaluate the proposed approach on two challenging datasets, E-VQA~\cite{evqa} and InfoSeek~\cite{infoseek}.
Experimental results show that the learned policy selectively applies region cropping when beneficial. Most notably, Region-R1 improves \emph{top-1 recall} over prior re-rankers in our setting. 
Ours CondRecall@1 improves 20\% on E-VQA and 8\% on InfoSeek. These gains translate into more reliable downstream grounding because MM-RAG systems often condition generation on the top-ranked evidence.

In summary, this paper makes the following contributions:
\begin{itemize}
\item We introduce \emph{query-side region cropping} for multi-modal re-ranking, and formulate it as a \emph{decision problem} to improve ranking.
\item We propose \textbf{Region-R1}, a \emph{candidate-agnostic} reinforcement learning framework that learns a region cropping policy optimized directly for multi-modal re-ranking.
\item Our experiment shows Region-R1 consistently improves performance on two challenging datasets, boosting CondRecall@1 by 20\% on E-VQA and 8\% on InfoSeek. We further provide diagnostic analyses of when region cropping is applied and ablations isolating the effect of each reward component.
\end{itemize}

\section{Related Works}

\paragraph{Knowledge-Based VQA.}
Knowledge-based Visual Question Answering (KB-VQA) requires models to answer questions by retrieving external knowledge beyond the query input~\cite{okvqa, aokvqa}. Early methods leveraged structured knowledge graphs~\cite{out-of-the-box, mucko, conceptbert, boosting, wu2022multi}, but often struggled with the open-ended and diverse nature of real-world questions. Recent work has largely shifted to Retrieval-Augmented Generation (RAG), retrieving evidence from unstructured corpora (e.g., Wikipedia). Benchmarks such as InfoSeek~\cite{infoseek} and Encyclopedic-VQA (E-VQA)~\cite{evqa} formalize this paradigm and emphasize fine-grained entity linking and multi-hop reasoning. State-of-the-art systems commonly follow a ``Retriever-Reranker-Generator'' pipeline. Wiki-LLaVA~\cite{wiki-llava} applies hierarchical retrieval, while EchoSight~\cite{echosight} and OMGM~\cite{omgm} use Q-Former-based re-ranking after initial retrieval.

More recently, studies optimize the interaction between VLMs and retrieval. ReflectiVA~\cite{reflectiVA} and MMKB-RAG~\cite{mmkb-rag} add self-reflective signals and consistency checks to decide when retrieval is needed, and ReAG~\cite{reag} and Iterative-RAG~\cite{iter-rag} investigate iterative retrieval to refine evidence chains. A persistent challenge is the semantic gap between visual queries and textual documents. To mitigate it, several works adapt query expansion from text retrieval~\cite{convgqr}, using generated captions or entity-aware rewriting to better guide search~\cite{adjali2024multi,zhu2024enhancing}. However, most still treat the visual input as a fixed global representation, underutilizing discriminative visual query reformulation for filtering distractors.

\paragraph{Multi-Modal Re-Ranking.}
Re-ranking is a critical refinement stage in MM-RAG pipeline, designed to re-order candidate evidence to maximize the rank of true positives. Recent approaches fall into two primary categories. \textit{(1) Interaction-Centric Methods.} These models rely on deep cross-modal alignment or generative reasoning to score candidates. EchoSight~\cite{echosight}, OMGM~\cite{omgm}, and ReflectiVA~\cite{reflectiVA} orchestrate fine-grained evidence fusion and self-reflective tokens to weigh relevance dynamically. RAMQA~\cite{ramqa} and Mario~\cite{ramezan2025multi} employ Large Multimodal Models as generative listwise re-rankers, synthesizing text and visual history to predict the optimal permutation of documents. \textit{(2) Representation-Centric Methods.} Alternatively, other works optimize the embedding space itself. EchoSight~\cite{echosight}, OMGM~\cite{omgm}, MM-Embed~\cite{lin2024mm}, and DocReRank~\cite{docrerank} force retrievers to distinguish subtle visual disagreements with hard negative samples.

Despite these advancements, both paradigms typically treat the visual query as a fixed, global tensor. Our work addresses this by introducing \textit{discriminative region cropping}, effectively reformulating the visual query for the re-ranking stage.
\begin{table*}[ht]
\centering
\small
\setlength{\tabcolsep}{5pt}
\renewcommand{\arraystretch}{1.10}
\begin{tabular}{lrrrrrr rrrrrr}
\toprule\toprule
\multirow{2}{*}{\textbf{Method}} &
\multicolumn{6}{c}{\textbf{E-VQA}} &
\multicolumn{6}{c}{\textbf{InfoSeek}} \\
\cmidrule(lr){2-7}\cmidrule(lr){8-13}
& \textbf{MRR} & \textbf{NDCG} & \textbf{R@1} & \textbf{R@5} & \textbf{R@10} & \textbf{R@20}
& \textbf{MRR} & \textbf{NDCG} & \textbf{R@1} & \textbf{R@5} & \textbf{R@10} & \textbf{R@20} \\
\midrule

EVA-CLIP
& 0.224 & 0.289 & 14.2 & 33.4 & 47.4 & 50.8
& 0.553 & 0.614 & 46.3 & 71.2 & 77.3 & 81.7 \\
ReflectiVA
& - & - & 15.6 & 36.1 & - & 49.8
& - & - & 56.1 & 77.6 & - & \textbf{86.4} \\
Wiki-LLaVA
& - & - & 3.3 & - & 9.9 & 13.2
& - & - & 36.9 & - & 66.1 & 77.9 \\
mR$^2$AG
& - & - & - & - & - & -
& - & - & 38.0 & - & 65.0 & 71.0 \\
\midrule

Random
& 0.060 & 0.160 & 1.2 & 5.9 & 20.7 & 50.8
& 0.193 & 0.323 & 10.6 & 27.1 & 40.7 & 81.7 \\
Center
& 0.143 & 0.220 & 7.5 & 20.1 & 31.5 & 50.8
& 0.468 & 0.545 & 38.8 & 56.8 & 66.6 & 81.7 \\
\midrule

Qwen2.5-3B
& 0.231 & 0.293 & 15.3 & 34.1 & 44.3 & 50.8
& 0.582 & 0.618 & 54.1 & 68.8 & 74.0 & 81.7 \\
Qwen2.5-7B
& 0.240 & 0.300 & 16.1 & 35.2 & 45.6 & 50.8
& 0.598 & 0.638 & 54.8 & 70.5 & 75.1 & 81.7 \\
\midrule

EchoSight
& \underline{0.402} & 0.423 & 36.5 & 47.9 & 48.8 & 48.8
& 0.586 & 0.631 & 53.2 & 74.0 & 77.4 & 77.9 \\
OMGM
& \textbf{0.473} & \textbf{0.500} & \underline{42.8} & \textbf{55.7} & \textbf{58.1} & \textbf{58.7}
& \underline{0.681} & \underline{0.717} & \underline{64.0} & \textbf{80.8} & \textbf{83.6} & 84.8 \\
\midrule

\textbf{Ours}
& \textbf{0.473} & \underline{0.480} & \textbf{44.7} & \underline{48.2} & \underline{49.9} & \underline{50.8}
& \textbf{0.706} & \textbf{0.732} & \textbf{66.5} & \underline{78.2} & \underline{80.2} & \underline{81.7} \\
\bottomrule\bottomrule
\end{tabular}

\caption{Re-ranking results on E-VQA and InfoSeek with top-20 candidates. We report MRR, NDCG, and Recall@$\{1,5,10,20\}$. Best in bold, second-best underlined.} 
\label{tab:main_results}
\end{table*}

\begin{table*}[h]
\centering
\small
\setlength{\tabcolsep}{5pt}
\renewcommand{\arraystretch}{1.12}
\begin{tabular}{l|ccc|ccc}
\toprule\toprule
\multirow{2}{*}{\textbf{Methods}} &
\multicolumn{3}{c|}{\textbf{E-VQA}} &
\multicolumn{3}{c}{\textbf{InfoSeek}} \\
\cmidrule(lr){2-4}\cmidrule(lr){5-7}
& \textbf{CondR@1} & \textbf{CondR@5} & \textbf{CondR@10}
& \textbf{CondR@1} & \textbf{CondR@5} & \textbf{CondR@10} \\
\midrule

EVA-CLIP~\cite{evaclip}      & 0.28 & 0.66 & 0.93 & 0.57 & 0.87 & 0.95 \\
ReflectiVA~\cite{reflectiVA}    & 0.31 & 0.72 & - & 0.65 & 0.90 & - \\
Wiki-LLaVA~\cite{wiki-llava}      & 0.25 & - & 0.75 & 0.47 & - & 0.85 \\
mR$^2$AG~\cite{mR2AG}     & - & - & - & 0.54 & - & 0.92 \\
\midrule

Random         & 0.02 & 0.12 & 0.41 & 0.13 & 0.33 & 0.49 \\
Center        & 0.15 & 0.40 & 0.62 & 0.47 & 0.70 & 0.82 \\
\midrule

Qwen2.5-3B~\cite{qwen2.5}  & 0.30 & 0.67 & 0.87 & 0.69 & 0.84 & 0.91 \\
Qwen2.5-7B~\cite{qwen2.5}  & 0.32 & 0.69 & 0.90 & 0.70 & 0.86 & 0.92 \\
\midrule

EchoSight~\cite{echosight}  & \underline{0.75} & \textbf{0.98} & \textbf{1.00} & 0.68 & \underline{0.95} & \textbf{0.99} \\
OMGM~\cite{omgm}       & 0.73 & \underline{0.95} & \underline{0.99} & \underline{0.75} & \underline{0.95} & \textbf{0.99} \\
\midrule

\textbf{Ours}   & \textbf{0.90} & \underline{0.95} & 0.98 & \textbf{0.81} & \textbf{0.96} & \underline{0.98} \\
\bottomrule\bottomrule
\end{tabular}

\caption{\textbf{Conditional recall} on E-VQA and InfoSeek. CondR@$K$ denotes CondRecall@$K$ (Eq.~\ref{eq:cond_recall}). Best in bold, second-best underlined.} 
\label{tab:condrecall}
\end{table*}

\section{Methodology}
\label{sec:method}

\subsection{Problem Definition}

We consider the re-ranking stage of a MM-RAG pipeline.
Each query is represented by an image-question pair
\(
x = (I_q, q)
\),
where \(I_q\) denotes the query image and \(q\) denotes the associated question.
Given a query, an upstream retriever produces a candidate set
\(
\mathcal{C} = \{c_j\}_{j=1}^{N}
\),
where each candidate \(c_j = (I_j, t_j)\) consists of an image \(I_j\) and textual description \(t_j\).
The objective of re-ranking is to produce an ordering \(\pi\) over \(\mathcal{C}\) that reflects relevance to the query.
Among candidates $\mathcal{C}$, the relevance labels $y \in \mathbb{R}^{N}$ are given. $y_j = 1$ for the positive candidate, and $y_j = 0$ otherwise.

Throughout this work, the candidate set is assumed to be fixed, and only the query representation is modified during re-ranking.

\subsection{Query-Side Region Cropping}

Let $d \in \{\textsf{REGION}, \textsf{FULL}\}$ denote a discrete decision variable indicating whether a region is selected from the query image.
When $d = \textsf{REGION}$, a continuous bounding box
\(
b = (x_1, y_1, x_2, y_2)
\)
is predicted.
A deterministic cropping operator \(g(\cdot)\) that extracts the region specified by \(b\).

The transformed query image $\tilde I_q$ is defined as

\begin{equation}
\tilde I_q =
\begin{cases}
g(I_q, b), & d=\textsf{REGION}, \\
I_q, & d=\textsf{FULL}.
\end{cases}
\end{equation}

The decision \(d\) and bounding box \(b\) are generated by a vision-language model conditioned on the query image and question.

\subsection{Scoring Model}
\label{sec:method:scoring}
Images and text are embedded into a shared representation space using a pre-trained vision-language encoder.
Let \(f_I(\cdot)\) and \(f_T(\cdot)\) denote the image and text encoders, respectively.

The query embedding is computed as
\begin{equation}
\mathbf{v}_q(d,b) =
\frac{
f_I(\tilde I_q)
}{
\big\| f_I(\tilde I_q) \big\|
}.
\end{equation}

For each candidate \(c_j \in \mathcal{C}\), the candidate embedding is defined as
\begin{equation}
\begin{aligned}
\mathbf{v}_j &=
\frac{f_I(I_j) + [t_j\neq\emptyset]\; f_T(t_j)}{\Big\|
f_I(I_j) + [t_j\neq\emptyset]\; f_T(t_j)
\Big\|}
\end{aligned}
\end{equation}

Candidate relevance scores are computed using cosine similarity
\begin{equation}
s_j(d,b) =
\cos\!\big(
\mathbf{v}_q(d,b),\,
\mathbf{v}_j
\big).
\end{equation}
Sorting \(\{s_j(d,b)\}_{j=1}^{N}\) in descending order yields a ranking \(\pi(d,b)\).

\subsection{Policy Learning and Reward Design}
\label{sec:policy_reward}

We fine-tune the vision-language model as a stochastic policy $\pi_\theta(a\mid x)$ that outputs a region cropping action
$a=(d,b)$ conditioned on the query $x=(I_q,q)$.
The discrete decision is $d\in\{\textsf{REGION},\textsf{FULL}\}$; when $d=\textsf{REGION}$, the model additionally predicts
a bounding box $b=(x_1,y_1,x_2,y_2)$.
Given $a$, we compute the ranking $\pi(d,b)$ using the fixed scoring model in Sec.~\ref{sec:method:scoring}.

\paragraph{Reward from re-ranking improvement.}
To explicitly optimize re-ranking quality, we define rewards using \emph{improvements} over the full-image baseline.
Let $\pi_{\textsf{full}}=\pi(\textsf{FULL},\cdot)$ be the baseline ranking and $\pi_a=\pi(d,b)$ be the ranking induced by action $a$.
We compute MRR and NDCG over the candidate pool, and define the corresponding deltas:
\begin{equation}
\begin{aligned}
\Delta \mathrm{MRR}  &= \mathrm{MRR}(\pi_a)-\mathrm{MRR}(\pi_{\textsf{full}}),\\
\Delta \mathrm{NDCG} &= \mathrm{NDCG}(\pi_a)-\mathrm{NDCG}(\pi_{\textsf{full}}).
\end{aligned}
\end{equation}

To further stabilize training, we include two auxiliary improvement signals.
Let $\mathrm{rank}(\pi)=\min\{r:\,y_{\pi(r)}=1\}$ denote the rank of the positive candidate.
Let $\mathrm{pos}=\max_{j:y_j=1}s_j$ and $\mathrm{neg}=\max_{j:y_j=0}s_j$ be the scoring of positive and negative candidates with their maximum score, and define $\mathrm{margin}(\pi)=\mathrm{pos}-\mathrm{neg}$. We then define:
\begin{equation}
\begin{aligned}
\Delta \mathrm{Rank} &=
\log \frac{\mathrm{rank}(\pi_{\textsf{full}})+1}{\mathrm{rank}(\pi_a)+1}, \\
\Delta \mathrm{Margin} &= \mathrm{margin}(\pi_a)-\mathrm{margin}(\pi_{\textsf{full}}),
\end{aligned}
\end{equation}
where the margin improvement term to provide a more direct training signal for \emph{discriminability} in the similarity space.
Ranking metrics such as MRR and NDCG depend on the induced ordering and can be relatively insensitive when many candidates receive similar scores or when small score changes do not alter the rank.
In contrast, maximizing $\Delta \mathrm{Margin}$ explicitly encourages the policy to increase the similarity between the query representation and the positive candidate while reducing the similarity to the most competitive negative candidates.

Eventually, when $d=\textsf{REGION}$, the reward is a weighted combination of these improvements:
\begin{equation}
\begin{aligned}
r(x,a)
&= w_1\,\Delta \mathrm{MRR} + w_2\,\Delta \mathrm{NDCG} \\
 &\quad + w_3\,\Delta \mathrm{Rank} + w_4\,\Delta \mathrm{Margin} \\
&\quad -\; \eta(b),
\end{aligned}
\label{eq:reward_region}
\end{equation}
where $\sum_{i=1}^{4} w_i = 1$ are scalar weights.
$\eta(b)$ penalizes malformed boxes (e.g., $x_2\le x_1$ or $y_2\le y_1$).

When $d=\textsf{FULL}$, we reward only if the baseline already ranks positive candidate at the rank 1, indicating the model makes the correct decision.
\begin{equation}
r(x,a)=\mathds{1}\!\left[\mathrm{rank}(\pi_{\textsf{full}})=1\right],
\qquad d=\textsf{FULL}.
\label{eq:reward_full}
\end{equation}

\paragraph{Optimization with r-GRPO.}

We optimize $\pi_\theta(a\mid x)$ using a region-aware variant of GRPO~\cite{grpo}, which we denote \textbf{r-GRPO}.
For each query $x$, we sample a group of $N$ actions $\{a^{(n)}\}_{n=1}^{N}\sim \pi_\theta(\cdot\mid x)$ and compute rewards $r^{(n)}=r(x,a^{(n)})$.
We form normalized advantages within the group
\begin{equation}
A^{(n)}=\frac{r^{(n)}-\mu_r}{\sigma_r+\epsilon}
\end{equation}
where $\mu_r$ and $\sigma_r$ are the mean and standard deviation of $\{r^{(n)}\}_{n=1}^{N}$.

A challenge in our setting is that the action space is structured as $a=(d,b)$, where $d$ is discrete while $b$ is continuous and only meaningful when $d=\textsf{REGION}$.
Naively applying GRPO can suffer from high variance and unstable updates when the sampled group is dominated by one decision.

To address this, r-GRPO uses decision-balanced group sampling.
For each query $x$, we sample a group of $N$ actions $\{a^{(n)}\}_{n=1}^{N}$ from $\pi_\theta(\cdot\mid x)$, while \emph{ensuring both decisions appear in the group.} This strategy prevents the more frequent decision from setting the baseline for the less frequent one, yielding lower-variance advantages. By balancing decisions in the sampled group and normalizing advantages conditional on $d$, r-GRPO stabilizes training for structured region-cropping actions while retaining the original GRPO update form.
\newcolumntype{T}[1]{>{\centering\arraybackslash}p{#1}}
\newcommand{\qimg}[1]{\includegraphics[valign=t,width=\linewidth]{#1}}

\begin{table*}[ht]
\centering
\setlength{\tabcolsep}{4pt}
\renewcommand{\arraystretch}{1.15}
\small
\begin{tabular}
{T{0.19\linewidth} T{0.19\linewidth} T{0.19\linewidth} T{0.19\linewidth} T{0.19\linewidth}}
\toprule
\makecell{\textbf{Query (Full)}\\\textit{+ question}} &
\makecell{\textbf{EchoSight}\\\textit{rank \#1}} &
\makecell{\textbf{OMGM}\\\textit{rank \#1}} &
\makecell{\textbf{Ours}} &
\makecell{\textbf{GT Positive}} \\
\midrule

\qimg{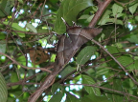} &
\qimg{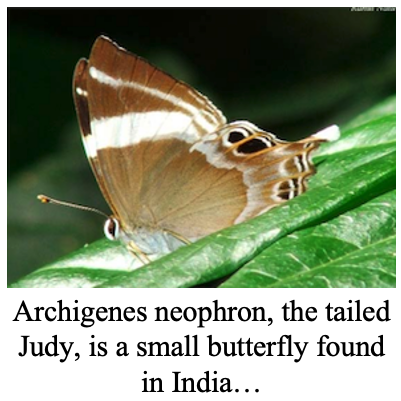} &
\qimg{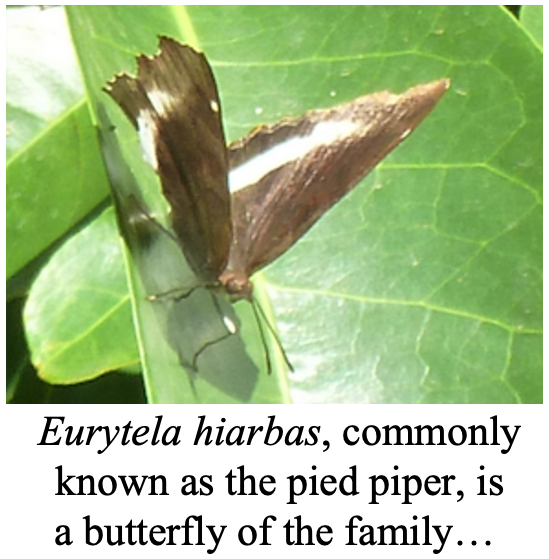} &
\qimg{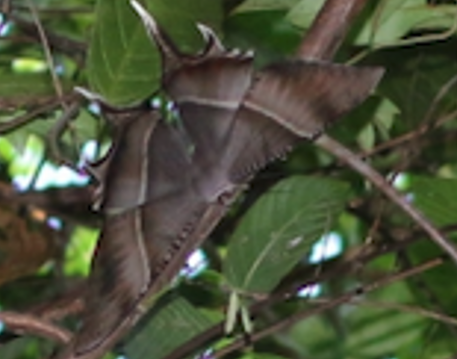} &
\qimg{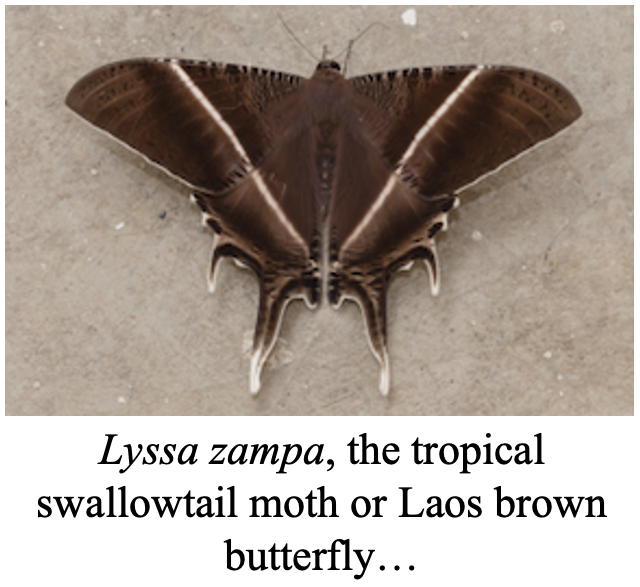} \\
\makecell[c]{\textit{\textbf{Q:} What is the closest}\\\textit{upper taxonomy of}\\\textit{this insect?}} &
\makecell[c]{\textcolor{red}{Wrong}} &
\makecell[c]{\textcolor{red}{Wrong}} &
\makecell[c]{\textcolor{blue}{Focuses on \textit{the insect}}} &
\makecell[c]{\textcolor{green!50!black}{Correct}} \\
\midrule

\qimg{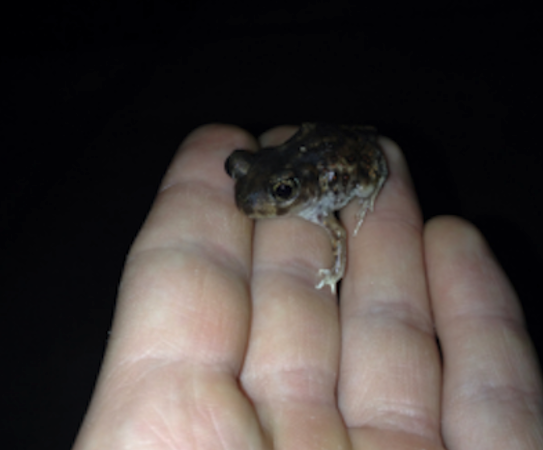} &
\qimg{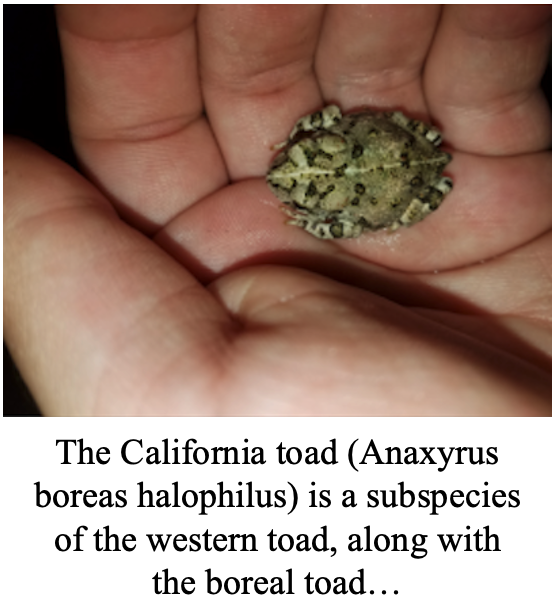} &
\qimg{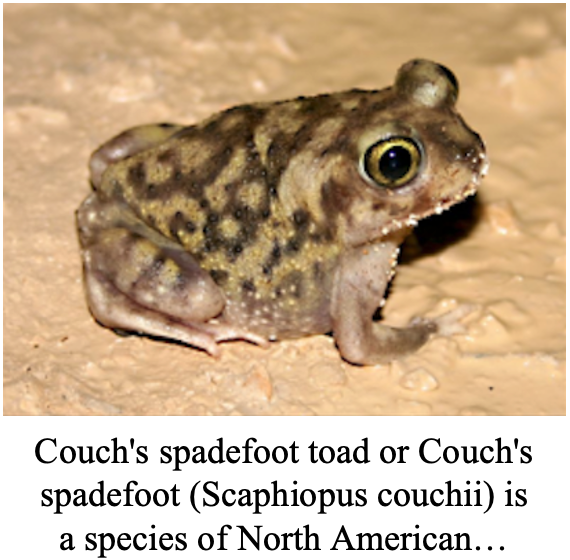} &
\qimg{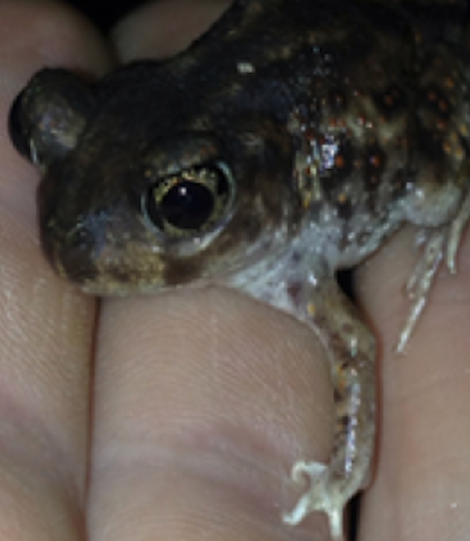} &
\qimg{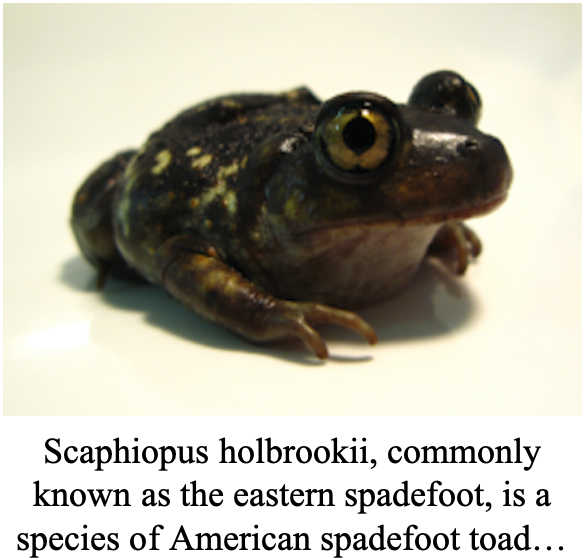} \\
\makecell[c]{\textit{\textbf{Q:} Where is this animal}\\\textit{native to?}} &
\makecell[c]{\textcolor{red}{Wrong}} &
\makecell[c]{\textcolor{red}{Wrong}} &
\makecell[c]{\textcolor{blue}{Focuses on \textit{the animal}}} &
\makecell[c]{\textcolor{green!50!black}{Correct}} \\

\bottomrule
\end{tabular}
\caption{Qualitative examples. Left-to-right: original query, top-1 from EchoSight~\cite{echosight} and OMGM~\cite{omgm} \textit{after re-ranking}, our query with region cropping, and the ground-truth positive candidate. Our region cropping flips the rank-1 prediction to the correct positive by removing distractors and emphasizing question-relevant region.}
\label{tab:qual_results}
\end{table*}

\section{Experiments}
\label{sec:experiments}

\subsection{Experimental Settings}

\paragraph{Datasets.} We conduct experiments on two challenging KB-VQA benchmarks, InfoSeek~\cite{infoseek} and E-VQA~\cite{evqa}, which are commonly used to evaluate MM-RAG systems.
Each dataset provides image-question pairs and a knowledge source containing multi-modal evidence, which are entity-centric Wikipedia content with associated images. We use Qwen-2.5-VL-3B~\cite{qwen2.5} as the base model for our experiments.

\paragraph{Training.}
During training, we construct candidate pools on-the-fly for each query by retrieving the top-$K$ candidates from the knowledge base using a pre-trained EVA-CLIP model.
Each candidate contains the evidence image and its associated text.
Following standard re-ranking practice, we retain only training queries for which at least one ground-truth relevant candidate appears in the retrieved top-$K$ pool, as re-ranking cannot recover missing positives. We set $K=20$ and retrieve the top-20 most relevant entities, balancing computational efficiency with retrieval effectiveness.

\paragraph{Testing and Evaluation.}
Following prior work~\cite{echosight,omgm}, we evaluate on the original test splits of InfoSeek and E-VQA.
For each test query, we retrieve a top-$K$ candidates from the knowledge base.
Re-ranking is then performed over this candidate pool.

We measure re-ranking performance using standard information retrieval metrics, including Mean Reciprocal Rank (MRR), Normalized Discounted Cumulative Gain (NDCG), and Recall@$K$.
Recall@$K$ reflects whether relevant candidate is ranked within the top-$K$ results and is particularly relevant for MM-RAG that condition downstream generation on a small number of retrieved candidates.
Since Recall@$K$ can be influenced by retrieval coverage, which may not directly reflect the re-ranking performance, we additionally report \emph{conditional} Recall@$K$ to better isolate re-ranking quality.
Let $\mathrm{hit}@K(i)\in\{0,1\}$ indicate whether query $i$ has positive candidate ranked within the top-$K$ positions after re-ranking.

\begin{equation}
\begin{aligned}
\mathrm{CondRecall}@K
&=\sum_{i\in\mathcal{Q}} \frac{\mathrm{hit}@K(i)}
        {\mathrm{hit}@20(i)} \\
&=\mathbb{E}\!\left[\mathrm{hit}@K \mid \mathrm{hit}@20=1\right].
\end{aligned}
\label{eq:cond_recall}
\end{equation}

where $\mathcal{Q}$ denotes the test query set.
Unless otherwise specified, same in training, we set $K=20$ and report Recall@$\{1,5,10\}$ together with CondRecall@$\{1,5,10\}$ and Recall@20. An ideal re-ranker that always promotes positive candidate to the top-$K$ whenever one is present in the top-$20$  achieves $\mathrm{CondRecall}@K = 1$ for all $K\le 20$.

\newlength{\cellH}
\setlength{\cellH}{0.19\textheight} 

\newcommand{\casecell}[1]{%
  \begin{minipage}[c][\cellH][c]{0.12\textwidth}
    \centering\bfseries #1
  \end{minipage}%
}

\newcommand{\imgcell}[2]{%
  \begin{minipage}[c][\cellH][c]{0.4\textwidth}
    \centering
    \includegraphics[height=0.93\cellH,keepaspectratio]{#2}\\[-2pt]
    {\footnotesize(\textbf{#1})}
  \end{minipage}%
}

\begin{table*}[ht]
\centering
\small
\setlength{\tabcolsep}{6pt}
\renewcommand{\arraystretch}{1.0}

\begin{tabular}{@{}c !{\vrule width 0.7pt} c c@{}}
\toprule 
\toprule
\textbf{Case} &
\makecell{\textbf{InfoSeek}} &
\makecell{\textbf{E-VQA}} \\
\midrule

\casecell{Original Rank = 1} &
\imgcell{a}{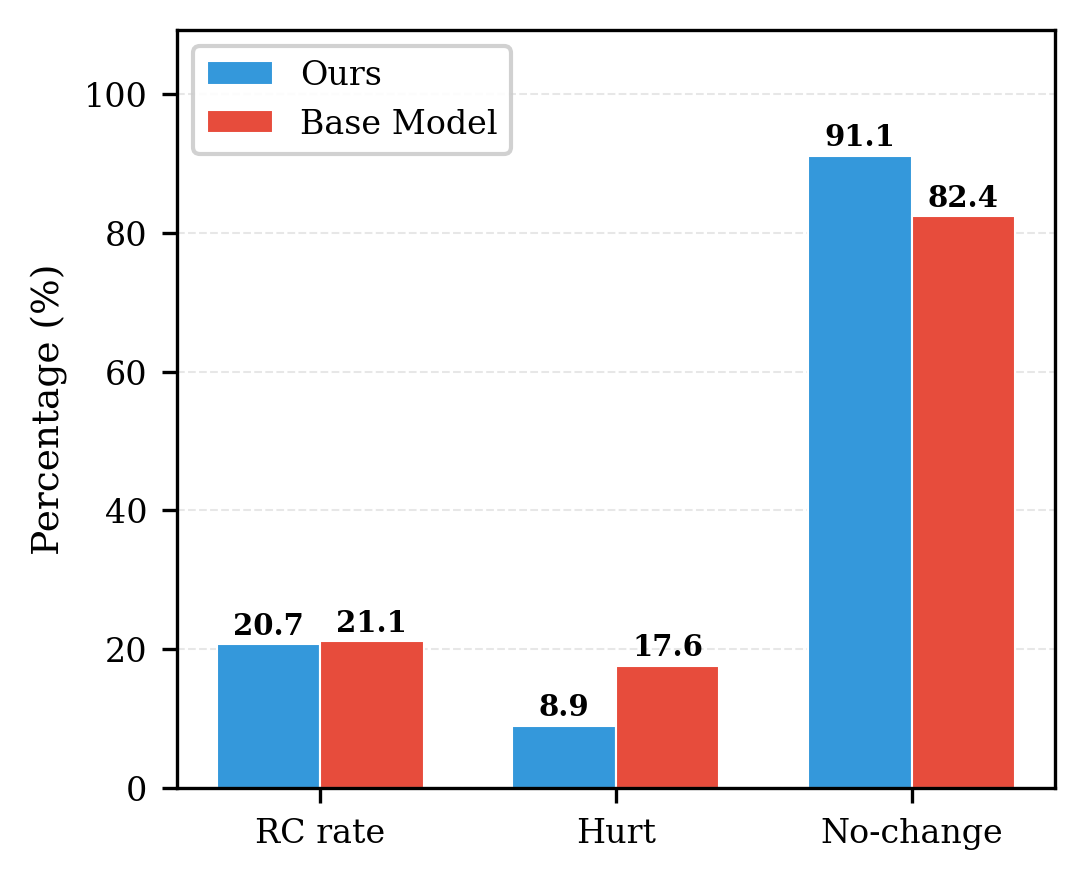} &
\imgcell{b}{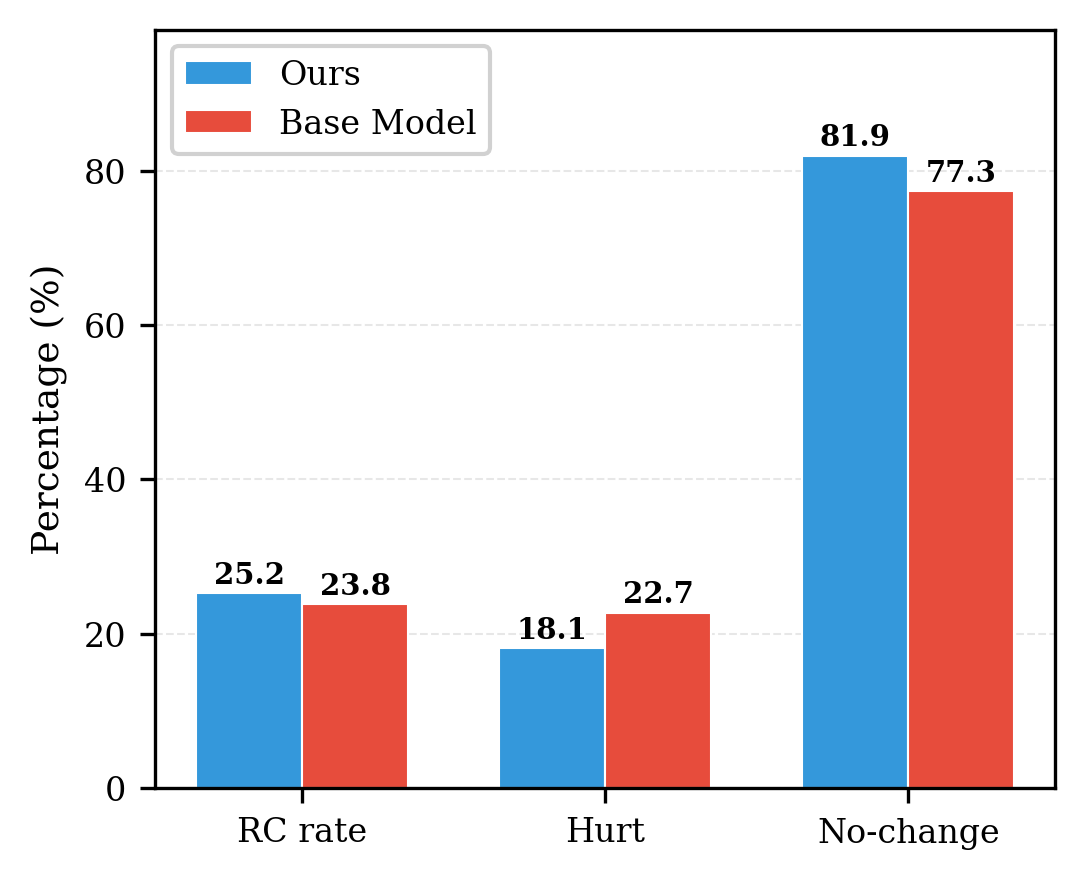} \\
\midrule

\casecell{Original Rank $> 1$} &
\imgcell{c}{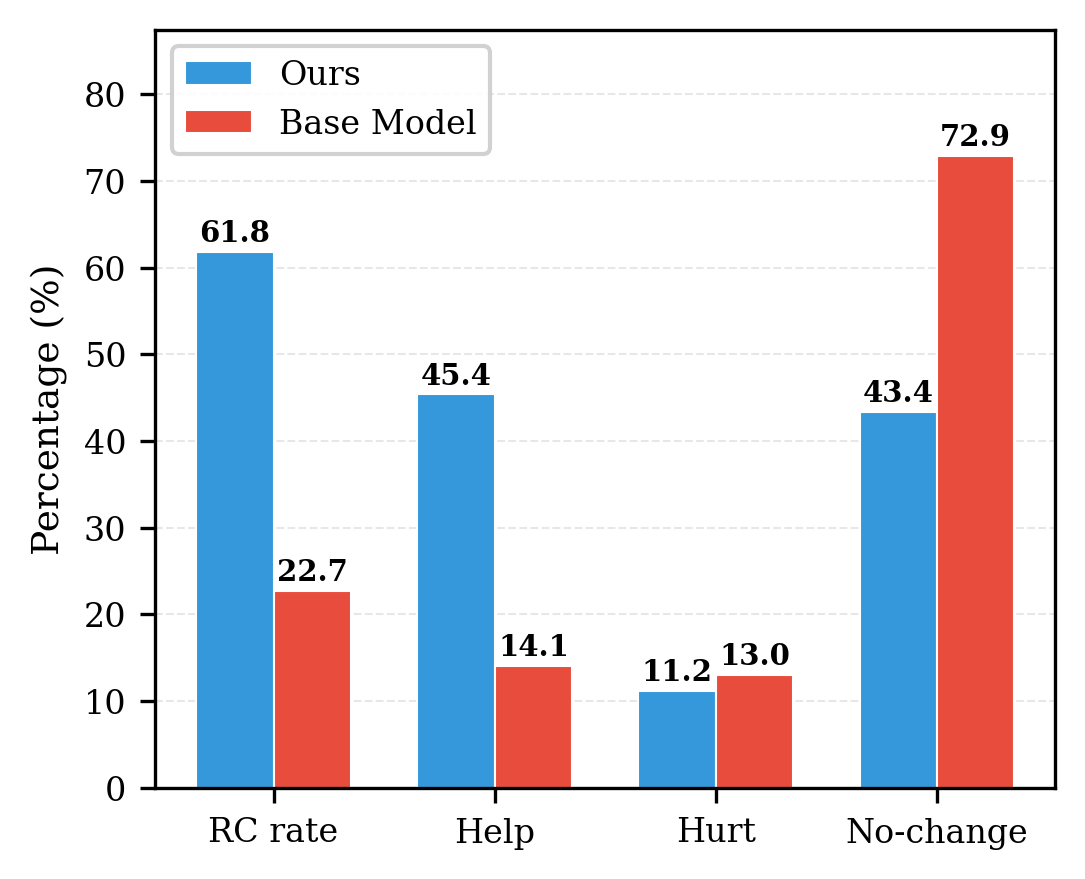} &
\imgcell{d}{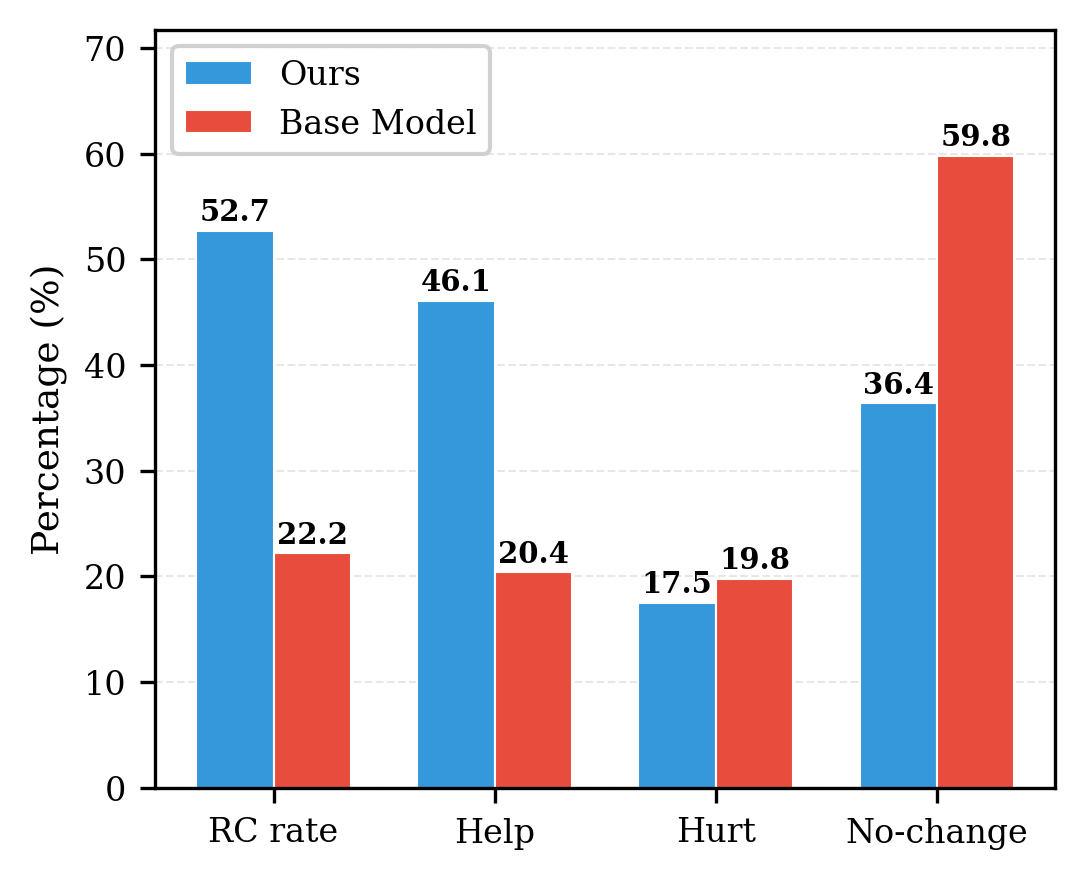} \\
\bottomrule
\bottomrule
\end{tabular}

\caption{\textbf{Quantitative results of our region cropping (RC) behavior across two datasets.}
(a,b) show cases where the original query already ranks the positive at \#1 on InfoSeek and E-VQA, respectively.
(c,d) show harder cases where the original rank is $>1$. \textit{RC rate} indicates how often does the model choose the crop a region.}
\label{tab:rs_behavior}
\end{table*}

\subsection{Baselines}
\label{sec:baselines}

We compare our query-side region cropping approach against baselines that reflect common design choices in multi-modal re-ranking.

\paragraph{Heuristic Region Cropping.}
We evaluate two non-learned region cropping strategies.
\emph{Center Region} selects a fixed central region of the query image.
\emph{Random Region} selects a region at a random location, with selected area matched to the distribution produced by the learned policy.

\paragraph{Zero-shot VLM Region Cropping.}
A vision-language model is prompted to predict a question-conditioned region of interest for the query image.
No fine-tuning or reinforcement learning is applied. In this study, we experiment with Qwen2.5-VL, with both 3B and 7B models.

\paragraph{Re-ranker Methods.}
We additionally compare against prior studies that optimizes the re-ranker trained with hard-negative samples, including EchoSight~\cite{echosight} and OMGM~\cite{omgm}, which improve multi-modal re-ranking by modifying the re-ranking model or query-candidate interaction mechanism rather than controlling the query representation.

\subsection{Main Results}
Tables~\ref{tab:main_results} and \ref{tab:condrecall} summarize retrieval/re-ranking performance across different approaches.
On InfoSeek, our learned \emph{query-side region cropping} consistently improves ranking quality over the \emph{No Region Cropping} baseline, with clear gains in top-heavy metrics.
Notably, the improvements concentrate on bringing the correct evidence to the very top of the list. Our method increases the frequency of ranking a positive candidate at rank \#1, indicating more effective ordering within the same candidate set.

Heuristic crops are brittle and can easily discard the evidence needed for matching. Random cropping performs poorly and a center crop only partially mitigates this effect.
In contrast, zero-shot VLM-based region cropping provides a stronger baseline than heuristics, but still falls short of the learned policy, A key reason is that the zero-shot VLM \emph{rarely commits to cropping}. Table~\ref{tab:rs_behavior} shows that the base model has a low RC rate, leaving the query unchanged for roughly 80\% of test examples, effectively behaving like the no-cropping baseline, suggesting that \emph{directly optimizing the cropping decisions for the re-ranking objective} is necessary.

Compared with prior retrieval/re-ranking methods, our approach achieves the strongest overall top-rank behavior with best MRR/NDCG and the highest Recall@1, while remaining competitive at larger cutoffs.
This trend is further supported by conditional recall in Table~\ref{tab:condrecall}. Our method attains the best CondR@1/CondR@5, confirming that when relevant evidence exists in the candidate pool, region cropping helps the model reliably surface it in the top positions.

Table~\ref{tab:qual_results} presents qualitative comparisons on different queries where the retrieved pool already contains the correct evidence, yet competing methods still place a distractor at rank \#1. In these cases, full image is dominated by irrelevant but visually salient regions, such as hands or leaves, causing the re-ranker to over-score mismatched candidates. By selecting a question-conditioned region before scoring, our method suppresses these distractors and shifts similarity toward the correct visual evidence to rank \#1.

\begin{figure*}[t]
\centering
\setlength{\tabcolsep}{6pt}
\renewcommand{\arraystretch}{1.0}
\begin{tabular}{cc}
\toprule
\textbf{Without margin term} & \textbf{With margin term} \\
\midrule
\includegraphics[width=0.38\textwidth]{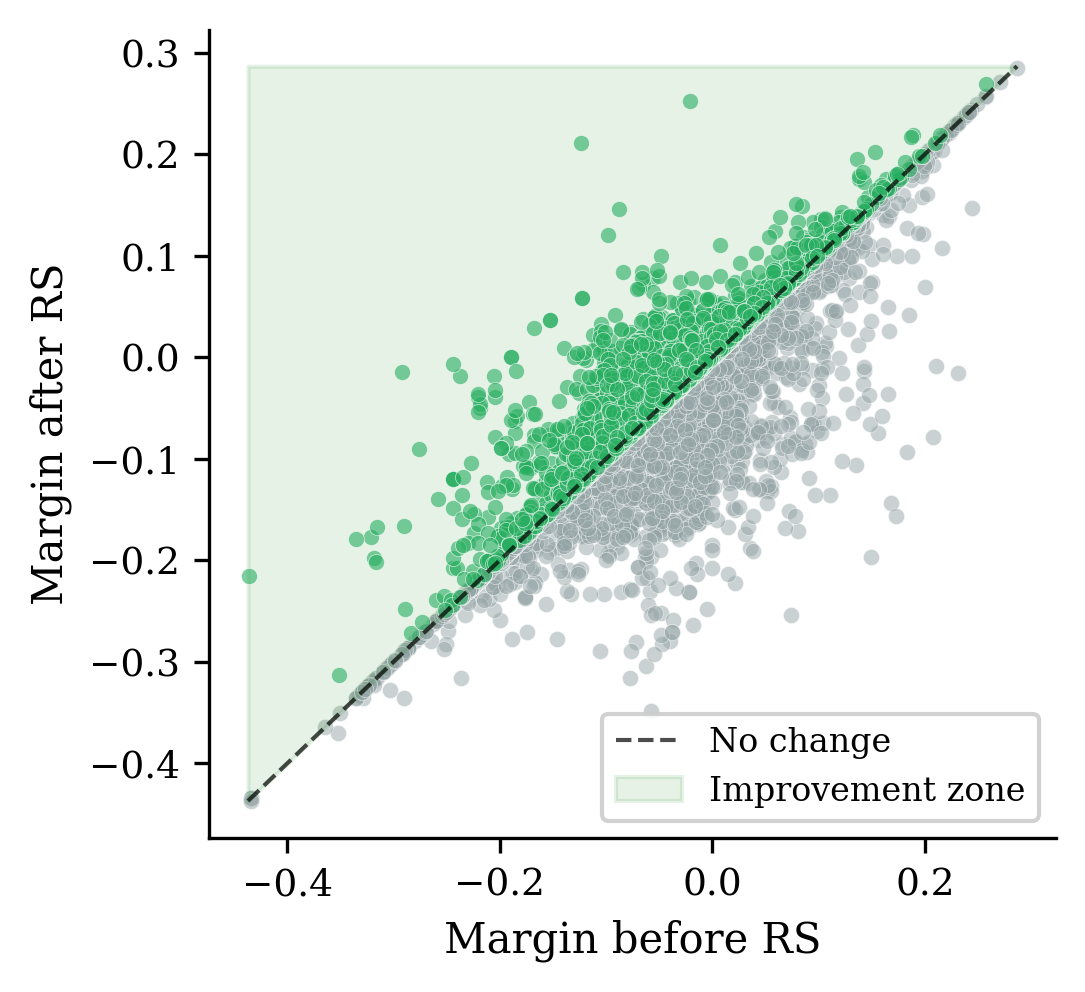} &
\includegraphics[width=0.38\textwidth]{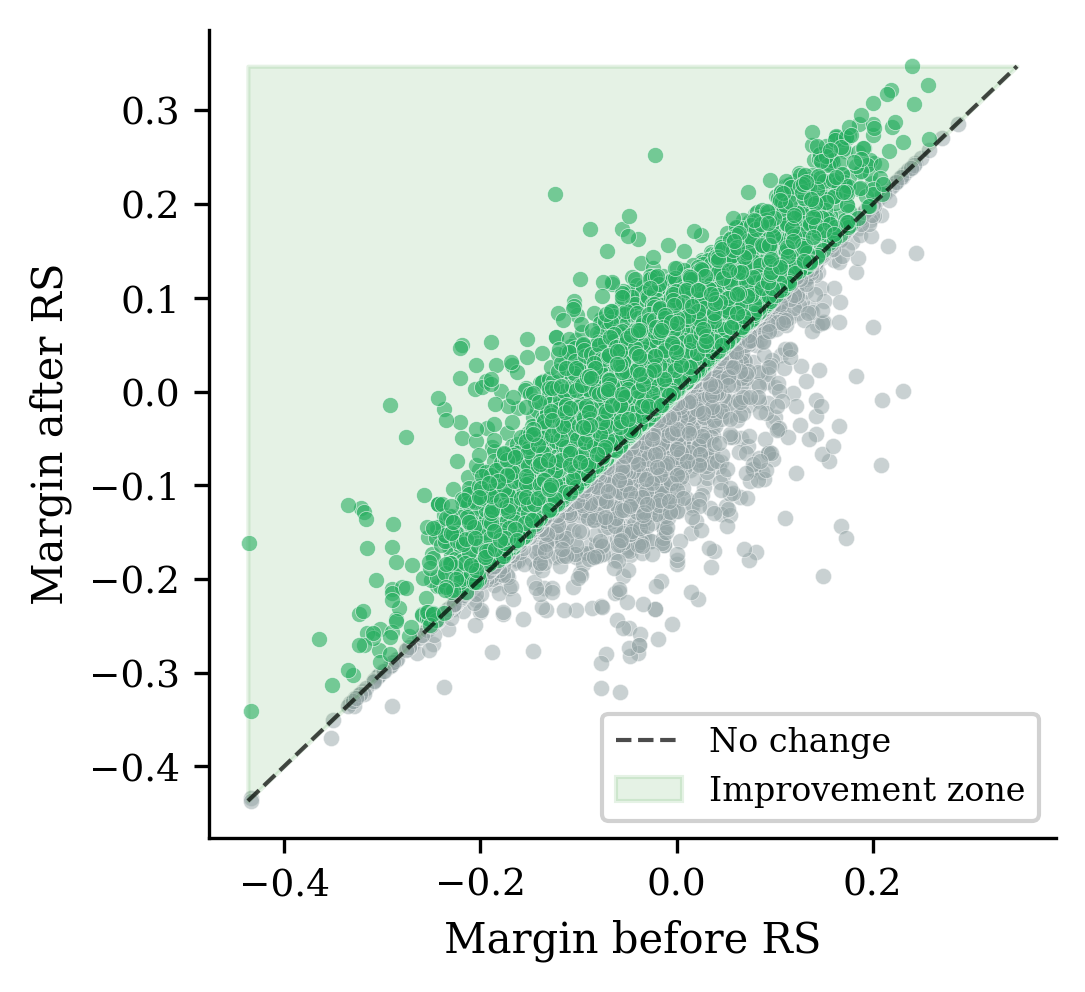} \\
\bottomrule
\end{tabular}
\caption{Ablation on the margin term in the reward design.}
\label{fig:margin_abl}
\end{figure*}

\subsection{Analysis}
\paragraph{Does the model perform region cropping (RC) correctly?}
Region cropping is not universally beneficial. Suppressing distractive regions can sharpen cross-modal similarity, but can also remove contextual cues that are necessary for disambiguation. Therefore, a well-behaved RS policy should (i) avoid altering the query when the full-image representation is already sufficient, and (ii) perform RS when the full-image representation is likely contaminated by distractors. To demonstrate whether our learned policy exhibits this capability, we analyze the RS behavior for two case: \textit{Original Rank = 1} and \textit{Original Rank > 1}. Table~\ref{tab:rs_behavior} presents how often the model chooses region cropping (\textit{RC rate}) and the resulting effect on the ranking, where \textit{Help}, \textit{Hurt}, and \textit{No-change} means the percentage of rank improves, degrades, and unchanged through test samples, correspondingly.

When the positive candidate is already at rank \#1, the model is expected to select region safely. Table~\ref{tab:rs_behavior}(a) and (b) show our learned policy does not simply disable RC, but it is less destructive than base model. Harmful outcomes decrease while \textit{No-change} remains dominant.

When the positive candidate is not at rank \#1, the model is expected to crop a region and suppress distractors. Table~\ref{tab:rs_behavior}(c) and (d) show our model performs RC substantially more often than the base model, yielding a higher \textit{Help} rate without increasing \textit{Hurt}. In contrast, the base model remains biased toward \textit{No-change}, which limits its ability to correct hard cases. Overall, Table~\ref{tab:rs_behavior} suggests that our learned policy acquires the capability of \emph{when to select region}.

\paragraph{Ablation study on reward function}
To isolate the contribution of components in the reward function, we train the policy under progressively richer reward variants while keeping the scoring model, candidate pool construction, and all hyperparameters fixed. Table~\ref{tab:reward_ablation} reports MRR on test sets of InfoSeek and E-VQA. From the table, ranking-metric deltas yield only marginal improvements, whereas introducing the margin term produces a huge jump. 

To further understand this effect, we analyze how RC changes the score separation between the true positive and the most competitive negative, i.e. the margin defined in Eq.~\ref{eq:reward_region}. Figure~\ref{fig:margin_abl} visualizes the margin before RC versus after RC. Without the margin term (left), most samples lie close to the no-change diagonal, with a substantial portion falling below it. In contrast, adding the margin term (Right) produces a clear shift. A much larger fraction of samples moves into the improvement zone above the diagonal, meaning that RC increasingly improves the margin rather than merely perturbing ranks. This supports the intuition that ranking-metric deltas provide sparse supervision, where small score changes may not flip an ordering. whereas $\Delta \mathrm{Margin}$ directly encourages the policy to pull the positive closer while pushing the hardest negative away. This study explains the substantial performance jump observed when the margin term is included.

\begin{table}
  \centering
  \small
  \begin{tabular}{lcc}
    \toprule
    \textbf{Reward Component} & \textbf{InfoSeek} & \textbf{E-VQA} \\
    \midrule
    $\Delta$MRR only
      & \texttt{0.611} & \texttt{0.408} \\
    + $\Delta$NDCG
      & \texttt{0.613} ($\uparrow$) & \texttt{0.425} ($\uparrow$) \\
    + $\Delta$Rank
      & \texttt{0.613} (-) & \texttt{0.426} ($\uparrow$) \\
    + $\Delta$Margin  (Full)
      & \textbf{0.706} & \textbf{0.473} \\
    \bottomrule
  \end{tabular}
  \caption{\textbf{Ablation on reward design.} Each row adds one component to the region-cropping reward (Eq.~\ref{eq:reward_region}).}
  \label{tab:reward_ablation}
\end{table}

\section{Conclusion}
Region-R1 demonstrates that re-ranking can be improved substantially by \emph{thinking the query image representation}, rather than only building heavier re-rankers. Our approach consistently outperforms prior multi-modal re-rankers across two challenging datasets, with particularly strong gains in retrieving the correct evidence at rank \#1. Future directions include adapting richer query, such as selecting multiple regions or learning soft, question-conditioned attention over the image. Another promising direction is to extend query-side adaptation from re-ranking to earlier retrieval stages under efficiency constraints, enabling end-to-end improvements while preserving scalibility.

\section{Limitations}
Region-R1 operates strictly at the re-ranking stage and therefore cannot recover evidence that is absent from the retriever’s top-$K$ candidate pool, so end performance remains bounded by retriever recall. In addition, our model is trained to optimize a fixed similarity-based objective, which may bias the learned region cropping behavior toward the scorer’s inductive biases and reduce transferability to other re-rankers or objectives. Furthermore, we evaluate on a limited set of benchmarks and a fixed re-ranking regime. While inference remains bounded by top-$K$, the method introduces additional decision/cropping overhead and RL-style training can be sensitive to reward design and hyperparameters.

\bibliography{custom}

\appendix
\section{Details of Dataset}
\label{app:dataset}
\textbf{Encyclopedic VQA}~\cite{evqa}
E-VQA is an encyclopedic visual question answering benchmark built around fine-grained entity recognition and evidence-grounded answering.
It contains approximately 221K question-answer pairs spanning 16.7K distinct entities, where each entity is associated with up to five images.
The entity inventory and images are sourced from iNaturalist 2021~\cite{van2021benchmarking} and Google Landmarks Dataset V2~\cite{weyand2020google}.
To support evidence-based reasoning, E-VQA additionally provides a controlled knowledge base derived from WikiWeb2M~\cite{burns2023wikiweb2m}, consisting of roughly 2M Wikipedia articles with images, from which supporting evidence for each answer can be retrieved.
Questions are categorized by reasoning complexity into single-hop and two-hop variants.
The dataset is organized into training/validation/test splits with approximately 1M, 13K, and 5.8K samples, respectively.
Following common practice for comparable evaluation, we restrict both training and testing to the single-hop subset.\\

\textbf{InfoSeek}~\cite{infoseek}
comprises 1.3M image-question-answer triplets grounded in visual entities, covering roughly 11K entities from OVEN~\cite{oven}. It includes 8.9K human-authored visual information-seeking questions alongside about 1.3M automatically generated questions.
The official split contains approximately 934K/73K/348K samples for train/validation/test, respectively.
Because the test split does not provide ground-truth answers, we report results on the validation set.
Notably, both validation and test emphasize generalization by featuring unseen entities and queries not observed during training.
InfoSeek is accompanied by a large knowledge base of about 6M Wikipedia entities. To align with prior work~\cite{echosight, omgm}, we adopt the same evaluation setting with a 100K-entity subset that still includes the 6,741 entities referenced by the training and validation questions.
Following this standard protocol, we evaluate on 71,335 validation samples.

\section{Implementation Details}
\label{app:impl}
We instantiate the region cropping policy from \texttt{Qwen2.5-VL-3B-Instruct}. Given a query image, the policy predicts either \textsf{Full} (keeping the original query representation) or \textsf{REGION}, which outputs a single 2D bounding box that specifies a region used to form an alternative query view for subsequent re-ranking.

For scoring and training feedback, we use \texttt{BAAI/EVA-CLIP-8B}~\cite{evaclip} to compute cosine-similarity scores between the query image and candidate evidence images, inducing the ranking used for both optimization and evaluation. When candidate-side text is available, we fuse it on the candidate side by combining candidate image and text embeddings before normalization. We train the policy with group-based RL using $N{=}8$ sampled outputs per query, learning rate $5e{-5}$ with a cosine schedule and $50$ warmup steps, batch size $4$ per device, and $2$ epochs; we cap the maximum prompt length at $8192$ tokens and generate up to $256$ new tokens. We adopt parameter-efficient tuning with LoRA applied to all linear layers of the policy backbone ($r{=}8$, $\alpha{=}32$, dropout $0.1$), updating only the LoRA parameters. We trained the model with 4 A6000 GPUs with 48GB VRAM. In addition, we provide the system prompt that we used to train the model in Table~\ref{tab:system_prompt}.

\section{Licenses}
The datasets we used, InfoSeek and E-VQA, are licensed under Apache License 2.0 and CC BY 4.0, respectively. The scoring model, EVA-CLIP-8B, is licensed under MIT License.
The prior re-ranker models from EchoSight and OMGM were released without an accompanying license. Qwen-2.5-VL series models are licensed under Apache License 2.0.

\clearpage
\onecolumn

\begin{longtable}{@{}p{\textwidth}@{}}
\caption{System Prompt}
\label{tab:system_prompt}\\
\toprule
\textbf{System Prompt} \\
\midrule
\endfirsthead

\toprule
\textbf{System Prompt (cont.)} \\
\midrule
\endhead

\bottomrule
\endlastfoot

\begin{lstlisting}[basicstyle=\ttfamily\footnotesize,breaklines=true,breakatwhitespace=false]
You are an intelligent region cropping assistant for re-ranking tasks.

Given an image and the user's question, your task is to decide whether finding a specific region would help remove the redundant information and rank more relevant information to a higher rank.

# Instructions
1. Carefully analyze the image in the context of the user's question.
2. Based on the user's question, decide whether region selection would improve re-ranking accuracy among many candidates, and push the most relevant candidate to the rank 1. You can decide:
   - If you think a specific region is most relevant to answering the question, select it to focus on it.
   - If you think the full image is already optimal for the question, output "FULL".
3. Output your decision with "FULL" or "REGION", if your decision is "REGION", you have to specify a region.

# Tools
<tools>
{"type":"function","function":{"name":"image_zoom_in_tool","description":"Zoom in on a specific region of an image.","parameters":{"type":"object","properties":{"bbox_2d":{"type":"array","items":{"type":"number"},"minItems":4,"maxItems":4,"description":"Bounding box as [x1, y1, x2, y2], where (x1, y1) is top-left and (x2, y2) is bottom-right."},"label":{"type":"string"}},"required":["bbox_2d"]}}}
</tools>

# How to call a tool
Return a json object with function name and arguments within <tool_call></tool_call> XML tags:
<tool_call>
{"name": <function-name>, "arguments": <args-json-object>}
</tool_call>

# Output format
{"Decision": "FULL" or "REGION", "Tool": <Call the tool with parameters if the decision is "REGION">}


Again, if you call the tool, you MUST follow the format exactly as specified. 

Otherwise, I will be unable to parse your response.
\end{lstlisting}

\end{longtable}

\clearpage
\twocolumn

\end{document}